**A Case Study to Reveal if an Area of Interest has a Trend in Ongoing Tweets**

**Using Word and Sentence Embeddings**

İsmail Aslan and Yücel Topçu

**Author Note**

İsmail Aslan linkedin.com/in/aslan-ismail/

Yücel Topçu linkedin.com/in/yucel-topcu/

We have no conflict of interest to disclose.

Correspondence concerning this article should be addressed to İsmail Aslan.

Email:ismailaslan1@gmail.com

## Abstract

In the field of Natural Language Processing, information extraction from texts has been the objective of many researchers for years. Many different techniques have been applied in order to reveal the opinion that a tweet might have, thus understanding the sentiment of the small writing up to 280 characters. Other than figuring out the sentiment of a tweet, a study can also focus on finding the correlation of the tweets with a certain area of interest, which constitutes the purpose of this study. In order to reveal if an area of interest has a trend in ongoing tweets, we have proposed an easily applicable automated methodology in which the Daily Mean Similarity Scores that show the similarity between the daily tweet corpus and the target words representing our area of interest is calculated by using a naïve correlation-based technique without training any Machine Learning Model. The Daily Mean Similarity Scores have mainly based on cosine similarity and word/sentence embeddings computed by Multilanguage Universal Sentence Encoder and showed the tweets' main opinion stream with respect to a certain area of interest, which proves that an ongoing trend of a specific subject on Twitter can easily be captured in almost real time by using the proposed methodology in this study. We have also compared the effectiveness of using word versus sentence embeddings while applying our methodology and realized that both give almost the same results, whereas using word embeddings requires less computational time than sentence embeddings, thus being more effective. This paper will start with an introduction followed by the background information about the basics, then continue with the explanation of the proposed methodology and later on finish by interpreting the results and concluding the findings.

## Introduction

The spread of mobile technology has recently made many people the users of easy-to-use social media platforms. Millions of people share and post millions of thoughts on a variety of subjects every day on different platforms. Therefore the number of people, and thus the online shares, is steadily increasing day by day. The amount and the availability of the data on social media has recently made social media analysis an important tool for social studies. For example, Ünver (2019) discussed in detail what programming, coding and internet research can do for International Relations Discipline and provided great examples of what can be done. He emphasized the importance of using Natural Language Processing (NLP) techniques to test more advanced theories of International Relations and claimed that there is a methodological gap in doing so. This methodological gap is a great venue in which NLP can find its place by providing many practical techniques to reveal information from text data, to characterize the textual documents, to follow ongoing trends, etc.

In social studies, data collection has traditionally been a difficult task and has been done through various techniques, such as interviews, questionnaires, surveys, oral histories, etc. All of these techniques are inherently applied in an artificial environment which may unintentionally end up with a bias in data collection. However, the social media platforms of today, which play an important role in our daily lives, provide a much more natural environment than these techniques to collect data to reveal the general attitude of people. Twitter is certainly the leading one of those platforms by an average of 6000 tweets generated every second that corresponds to 500 million tweets in a day and 350,000 tweets in a minute (Sharma & Daniels, 2020), thus providing very useful data to let us understand the public opinion about a certain phenomenon.

On Twitter people share their opinion in the natural flow of their lives and this natural sharing leads them to be more sincere about their opinion by minimizing the bias caused by the data collection techniques. People also have to share their opinion in 280 characters on Twitter, which leads them to succinctly share their thoughts, thus concentrating on the most important sentiments they want to emphasize. Therefore, in terms of providing data collected more naturally and succinctly, Twitter has a superiority over traditional data collection techniques.

Considering the data Twitter can provide, in any context, such as social or economic issues, being able to follow the fast-changing social attitude is of great importance for the decision makers. Tweeter is a very dynamic social network that allows people to instantly share their opinions every second on different subjects. To able to quickly analyse tweet text data and detect trending attitudes requires fast and easily applicable NLP techniques in order to extract information automatically since to make an overall inference about the public opinion by reading all those tweets is not possible for a human being taking into account the the speed of the information sharing, the amount of the people sharing information, the variety of the topics/opinions, etc. This has led to many studies conducted to extract information from Twitter text data such as sentiment analysis, opinion mining, etc., as mentioned by Giachanou and Crestani (2016); all focused on catching the sentiment on the tweets.

In this study, however, our focus is to reveal if an area of interest has a trend in ongoing tweets. In other words our purpose is to reveal if a given, particular sentiment is in the tweets as opposed to retrieving the general opinion or sentiment in the tweets as in aforementioned studies. In order to achieve this purpose, we used word/sentence similarity metrics calculated by using cosine similarity over Multilanguage Universal Sentence Encoder (MUSE) embeddings.

The objectives of the study are:

● To develop an easily applicable and automated method that uses already developed models without any extra model training and data preparation such as labeling data in order to follow trends in social media, namely Twitter text data.

● To detect the correlation between tweets and the target words representing the area of interest / target topic.

● To compare the effectiveness of sentence level embeddings and word level embeddings.

## Background

### Word Embeddings

In NLP problems, one big issue is the representation of a word in numbers in order to make quantitative analysis. Without having a practical and beneficial numerical representation of a word, it would not be possible to develop Machine Learning (ML) / Deep Learning (DL) models which require numbers/numerical data. An embedding for a word means a continuous vectorial representation of the given word in a predefined-dimensional space. Each word vector is trained to have a semantic relationship with another. (Mikolov et al. 2013; Pennington et al. 2014) It has been shown that word embeddings accurately capture the semantics and context of words and they can also be used in any kind of ML/DL models as a representation of a word (Le & Mikolov, 2014; Severyn & Moschitti, 2015; Socher et al., 2011).

Mikolov et al. (2013) proposed a model to create a real-valued vectorial representation for each word. With this model, it is shown that not only will the embeddings of similar words tend to be close to each other in the multi-dimensional space, but also the word embeddings can have multiple degrees of similarity. The very well-known example of this study is to apply simple algebraic operations to reach semantically similar words.   That is, vector("King") - vector("Man") + vector("Woman") results in a vector that is closest to the vectorial representation of the word "Queen". Another example is vector("Paris") – vector("France") + vector ("Italy") = vector("Rome"). Another model called GloVe also produced word embeddings that showed that the embeddings are a very meaningful representation of the words in a multi-dimensional space (Pennington et al., 2014).

**Sentence Embeddings**

Both models mentioned above produce word embeddings ignoring the sentence structure. However, recently developed models aim to create embeddings to represent the whole sentence instead of a single word. These models allow us to interpret the entire sentence giving a more meaningful embedding vector by taking into account the sentence as a whole, considering, for example, negations, phrasal verbs, etc. Yang et al. (2020) presented MUSE to compute embeddings for sentence-length texts.

**Multilanguage Universal Sentence Encoder**

MUSE, a pretrained model in TensorFlow library, creates embeddings for 16 different languages into a single shared semantic embedding space and achieves performance in cross-lingual semantic problems. In this study, however, we have used MUSE to compute embeddings for the English language only and did not use the cross-lingual function of the model. Additionally, even though MUSE is trained for sentence-length texts, we have also used it to produce word embeddings.

The example illustrated in Figure 1 clearly shows that similarities between the sentences can be captured by the MUSE model (TensorFlow, 2021). In this example, the semantic similarity of different sentences are computed as the inner product of the encodings. The darker the areas in the heatmap, the more correlated the sentences.

**Figure 1**

*Semantic Similarity of Sentences*

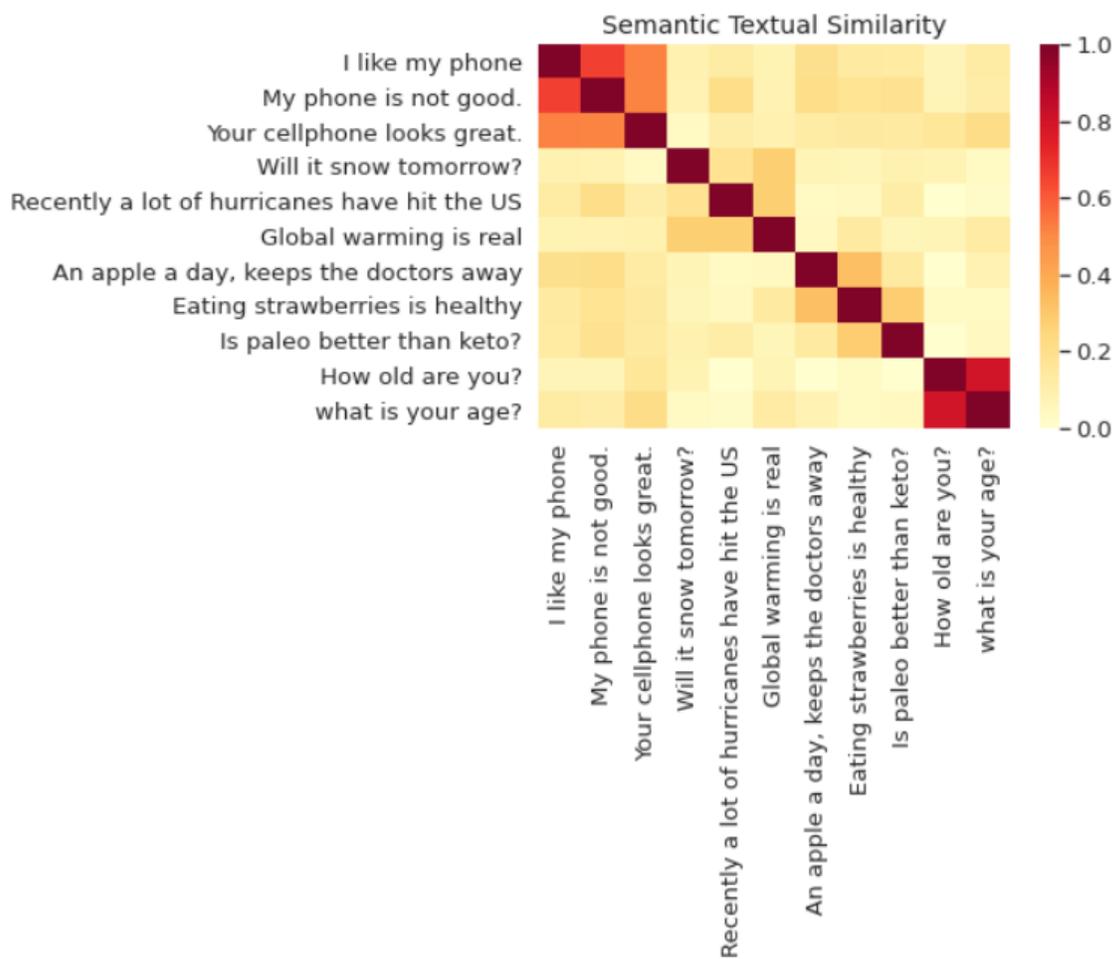

**Text Similarity Metrics**

In NLP studies there are different measures proposed and used for different NLP tasks

such as information retrieval (Li & Xu, 2014), automatic question answering (Jiang & de

Marneffe, 2019), machine translation (Wang et al., 2019), dialogue systems (Serban et al., 2016),

and document matching (Pham et al., 2015). The main idea behind similarity is very intuitive and it is to capture the commonness or semantic relatedness between words in a numerical way (Wang & Dong, 2020).

Any distance metric such as Euclidean, Cosine, Manhattan, etc. can be used to measure the similarity between word/sentence vectors whereas each vector represents a different word/sentence. However, when investigating the studies in this area, one can easily see that cosine similarity is more widely used than the other metrics such as in Yang et al. (2018) and Yang et al. (2019). We will also use cosine similarity in this study.

**Azerbaijan – Armenia Conflict**

We applied our methodology to the tweets posted during the conflict between Armenia and Azerbaijan in 2020. Admittedly, our upcoming plan is to do research in social science as the mentioned conflict being the case of our focus. This paper will serve as a base which explains the quantitative aspects of the entire study and which also finds out the most effective method to implement in our future study.

Even though there has been a long-running dispute over the Nagorno-Karabakh region between these two countries, it has been a frozen one for more than a decade. However, the tension increased in the summer of 2020 and turned into an armed conflict thereafter. Leaving the details of the conflict to the other paper, the main timeline which will be important for us to interpret the results are as follows (Topçu, 2021).

- In June 2020, the ongoing tension started to increase slightly.

- In July 2020, there happened a border issue causing the tension to increase seriously.

- The tension released for approximately two months thereafter.

- On 27 September 2020, the armed conflict started causing a regional war between two countries called the 44 Days War.

## Methodology

The main steps of our methodology are collecting tweets, cleaning the data, creating frequency tables, selecting target words, computing embeddings, calculating Daily Mean Similarity Scores and creating the plots. The entire process is shown in Figure 2 and the details are explained afterwards. We used two methods to calculate mean similarity scores:

Method 1: Calculation of mean similarity scores using word embeddings

Method 2: Calculation of mean similarity scores using sentence embeddings

The general framework of the methodology is the same for both methods except some minor changes detailed in Figures 2 and Figure 4 and in the related explanations.

**Figure 2**

*Methodology Framework*

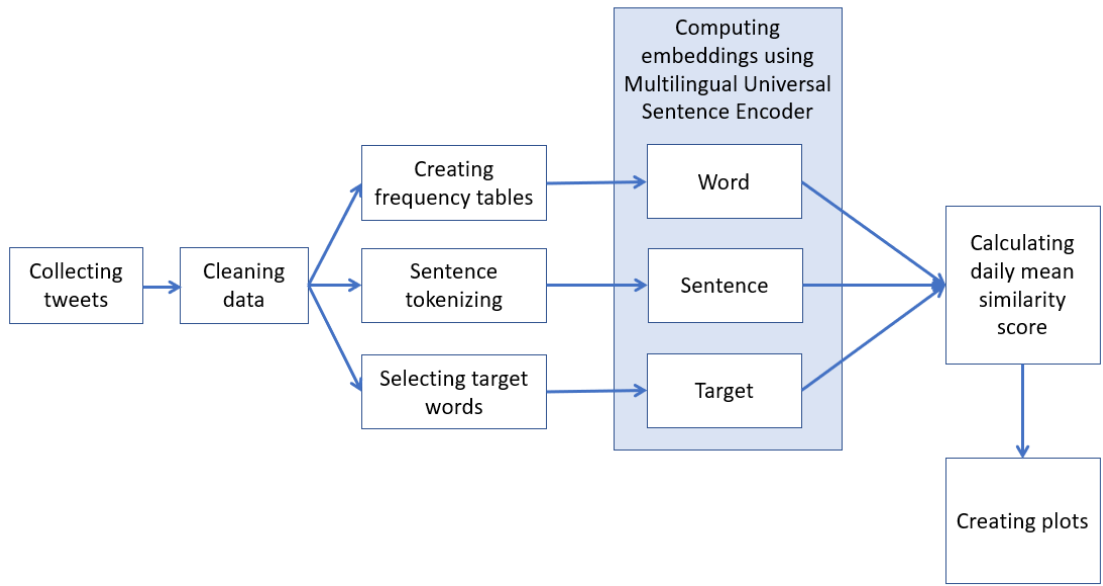

**Collecting Tweets**

We used Social Network Scraper (GitHub, 2021), a python based library, in order to collect the tweets. Even though we have Twitter API, we could not use it for tweet collection because it was not practical to collect tweets between certain dates in the past via Twitter API.

We collected the tweets posted worldwide in English language between 01 April 2020 and 31 March 2021 by having in the middle of this period the June - November 2020 timeframe which is our focus.

While collecting the tweets we used "Azerbaijan" as the keyword and set the language to English. We preferred to conduct our study on English tweets due to the lack of well-performing pre-trained models for Azerbaijani and Armenian, as well as we do not have any knowledge about these two languages. We did not collect retweets.

**Cleaning Data**

Any tweet content may include up to 280 characters long text, images, GIFs, and/or videos. Most users also have mentions, hashtags, web addresses, and/or URLs within the text. The fields other than text are not in the interest of the analysis and do not provide useful information for the purpose of this study since our purpose is related with the information/opinion mentioned in the text. Therefore, during the tweet collection phase, we only collected the text data excluding images, GIFs, videos, etc., and then, during cleaning, we removed all special strings mentioned above, as well as the emojis, from the text data.

**Creating Frequency Tables**

Since this step is not required for the second method based on sentence embeddings, we created frequency tables only for the first method which is based on word embeddings. In doing so, we used the Python WordCloud library  and generated the corpus of the day out of the collected tweets posted on that day, then removed the stopwords and created the frequency tables of the words for each day in the end.

**Selecting Target Words**

In order to reveal whether our area of interest has a trend in tweets, first we need to pick the words that represent our area of interest. We call these words target words. Since the area of interest we focused on in this study is related to a conflict, it can be represented by the words "war" and "peace". Even though these two words seem to have opposite meanings, they both come from the same domain, only the word "peace" not being as aggressive as the word "war". We also picked the word "computer" as a third word, which comes from a totally different domain that is not related to the conflict, international relations, etc. The purpose of selecting a word like "computer" from a completely different domain is to use it in a fashion similar to a control group in an empirical study.

In Figure 3 the similarity and the correlation can be clearly seen between the target words. The darker the areas, the more correlated the words with each other. The words from the same domain have higher correlation. One can observe on the heatmap that  the word "conflict" has a strong correlation with the words "war" and "peace"  and a weak correlation with the word "computer". Investigating the heatmap, it is easily understandable that the words "war" and "peace" are from the same semantic domain as the word "conflict" whereas the word "computer" is from a different domain. Therefore, we can use the words "war" and "peace" to represent our area of interest and the word "computer" as a control variable to check the reliability of our method, thus all three are being our target words.

**Figure 3**

*Semantic Similarity of Target Words*

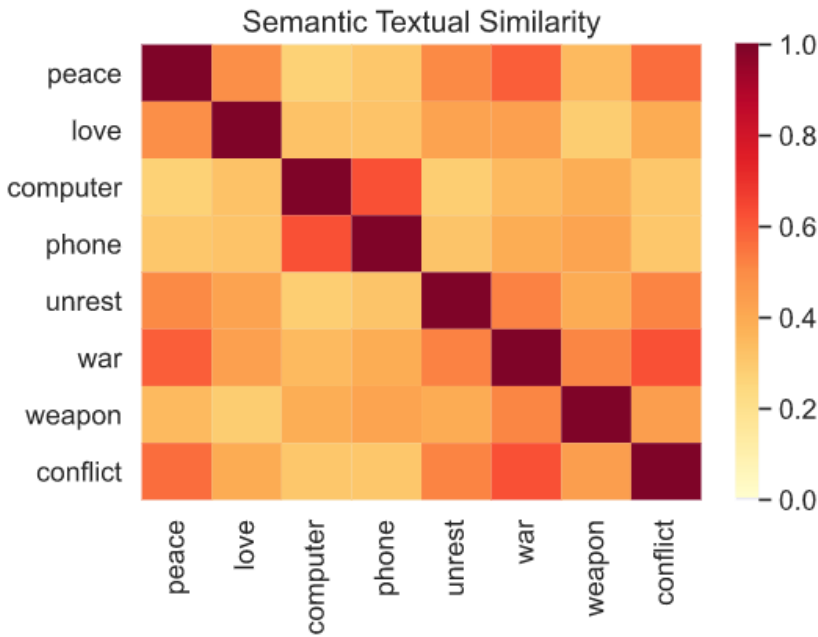

## Computing Embeddings

As mentioned before, we used MUSE to compute both word and sentence embeddings in this study. For the first method based on word embeddings, the fact that world clouds are widely used to get an overall idea about a corpus has led us to consider a calculation of similarity between the most frequent words and the target words. So, we computed the word embeddings for the top 1000 words in the daily word frequency tables, thus having 1000 word embeddings for each day during one year period.

For the second method based on sentence embeddings, we divided the daily tweet corpus into sentences using sentence tokenizer and removed the sentences less than 3 characters, then computed the sentence embeddings for each sentence.

**Calculating Daily Mean Similarity Scores**

We used cosine similarity to calculate the similarity between embeddings. The details of the calculation are illustrated in Figure 4.

For the first method based on word embeddings, in order to calculate Daily Mean Similarity Scores, we calculated the average of the similarity between each target word and each of the most frequent 1000 words taken from the frequency table of the given day. The result provided us with the mean similarity score of that day for a target word. Related pseudocode for a single target word is shown below.

```
mean similarity score = 0
total score = 0
for word in top n words:
        temp similarity = cosine similarity (word embedding, target word embedding)
        total score += temp similarity
mean similarity score = total score / n
```

For method 2 based on sentence embeddings, in order to calculate Daily Mean Similarity Scores, we calculated the average of the similarity between each target word and each sentence in daily tweet corpus. The result once again provided us with the mean similarity score of the given day for the given target word, based on sentence embeddings. Related pseudocode for a single target word is shown below.

```
mean similarity score = 0
total score = 0
for sentence in daily tweet corpus:
        temp similarity = cosine similarity (sentence embedding, target word embedding)
        total score += temp similarity
mean similarity score = total score / number of sentences in the daily tweet corpus
```

**Figure 4**

*Calculation of Daily Mean Similarity Scores*

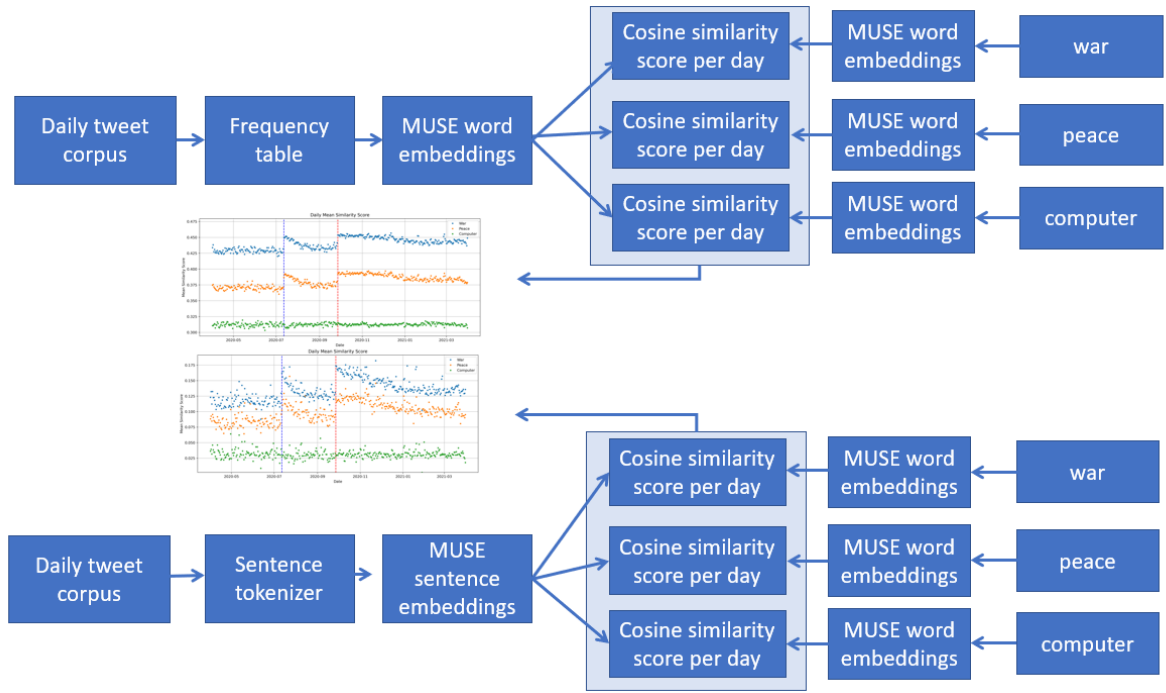

**Creating Plots**

      The Daily Mean Similarity Score gives us a strong indication of whether people are talking or not talking about a specific topic (target words), or they are completely irrelevant to the topic in a given time frame.

      For each method we created one plot illustrating the mean similarity scores of the target words for each day of the given period. The critical dates of the conflict are also drawn on the plots which will be interpreted in the next section.

## Analysis and Results

Totally 1,524,016 tweets which are posted in the period of 01 April 2020 and 31 March 2021 are collected . The daily tweet counts are shown in Figure 5. The peaks in the plot coincide with the conflict timeline mentioned before.

**Figure 5**

*Daily Tweet Counts*

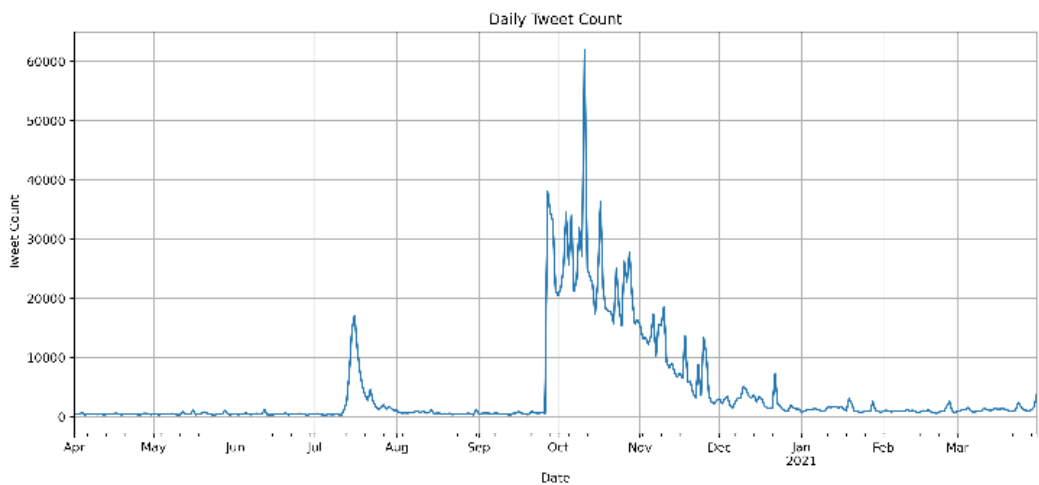

**Interpretation of Daily Mean Similarity Score Plots**

In the Daily Mean Similarity Score plot in Figure 6, which is produced based on word embeddings, the higher the mean similarity score of a target word, the higher the number of the words semantically similar to that target word in daily tweet corpus, which means that the topic represented by the given target word has a trend on Twitter.

**Figure 6**

*Daily Mean Similarity Scores of Word Embeddings*

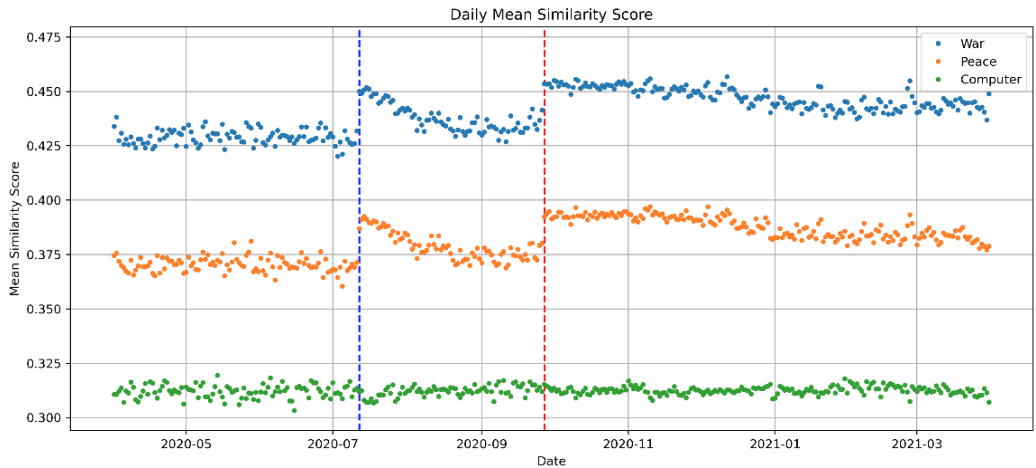

By looking at the plot, one can see that there is a high correlation between conflict timeline and the Daily Mean Similarity Scores. Namely, there are two picks in the semantic similarity plot. The first peak represented by the blue vertical dashed line corresponds to the date of the border issue which took place on 12 July 2020. The second peak represented by the red vertical dashed line corresponds to the date, 27 September 2020, on which armed conflict between the two countries started. After the second peak, when the conflict escalated to a full scale war, we observed a more condensed and higher level score compared to the score before the first peak, a time period during which there is no observable change in the similarity scores. We observe similar trends both for the word "war" and "peace" as we expected. On the other hand, the target word "computer" which is a control variable does not have any trend during the same period which proves that the trend in the words "war" and "peace" is not by chance but reliable.

In addition to these significant observations, there are also some relatively minor findings on the plot. On 10 November 2020, the President of Azerbaijan, the Prime Minister of Armenia and the President of Russia signed a statement declaring a complete ceasefire and end to all military operations in the conflict zone. The ceasefire was violated on 11 December 2020 but the incident was immediately restrained before causing an escalation in tension again and then a more stable situation was reached by mid-December. These events also explain the small peak and the later-on decrease in the mean similarity scores after December 2020.

It seems that the trends in the tweets we analyzed are just following the incident almost in real time. We do not observe any considerable change in the similarity scores before the first peak. However, we observe an increase in the similarity scores just before the second peak for both target words which may serve as an indication of upcoming war.

In the Daily Mean Similarity Score plot in Figure 7, which is produced based on sentence

embeddings, we observe similar patterns as in the previous plot illustrated in Figure 6, but the scores are more dispersed and not as stable as the similarity scores calculated based on word embeddings. There is no additional information we can reveal from the plot based on sentence embeddings other than what we obtained from the plot based on word embeddings.

**Figure 7**

*Daily Mean Similarity Scores of Sentence Embeddings*

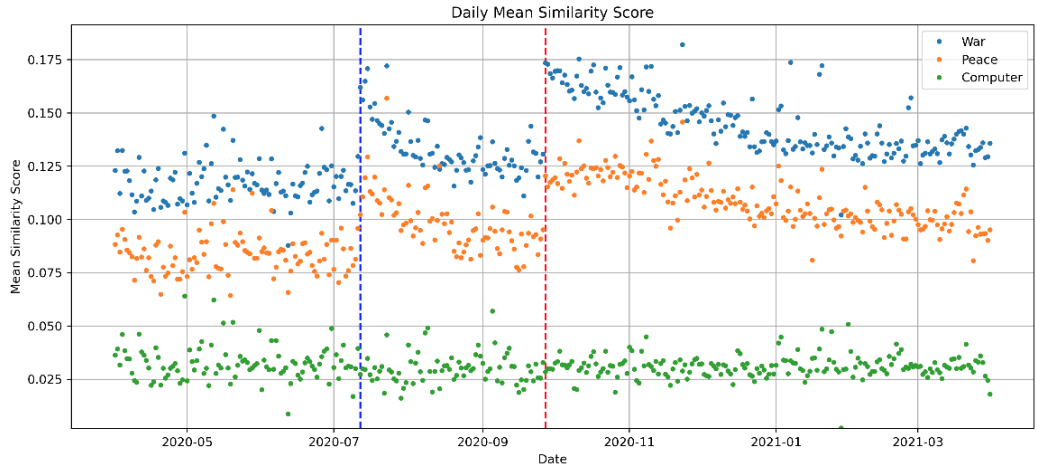

**Comparison of Two Methods**

The two different methods give almost the same results. The plot based on sentence embeddings are more dispersed while the method based on word embeddings show more distinguishable lines. That might be because in the method based on word embeddings we select the top most frequent words which might help reduce the noise in the tweet texts. On the other hand, sentence embeddings use all the sentences in the daily tweet corpus, thus having a wide semantic diversity.

From the computational cost aspect, the method based on sentence embeddings requires the computation of an embedding for each sentence in the daily tweet corpus. That increases the computational cost very much. On the other hand, counting and sorting daily word frequencies also creates a computational cost; however, these steps take far less time compared to the computation of sentence embeddings.

**Conclusion**

Word and sentence embeddings are very powerful representations used in NLP, which allow us to create ML/DL models for natural languages based on different requirements such as classification, named entity recognition, question answering, etc. There are, of course, the embeddings available for us to create models but a more important issue is how to make use of them in real life in order to benefit from the knowledge that using embeddings can provide. Having this idea in our mind, we developed an easily applicable automated methodology that uses already developed models without any extra model training and data preparation in order to follow trends in social  media, namely tweet text data, which is a tremendous task if done manually.

When we applied the proposed methodology to the collected tweets of this study, we observed that the trend of the target words "war" and "peace" coincides with the conflict timeline while our control target word "computer" does not have any change in its pattern during the course of the conflict as we expected. Therefore, the ongoing trend of a specific subject on Twitter can easily be captured in almost real time by using the proposed methodology in this study.

While applying the method, we used both word and sentence embeddings as an alternative to each other and compared the effectiveness of the two. We observed that while both provide the similar results, using sentence embeddings increases the computational cost a lot. Therefore, we recommend using word embeddings over sentence embeddings in the proposed methodology.

Even though not on purpose, this study also showed that MUSE can also be used to compute word embeddings.

This study is conducted only for the tweets in English so a future study conducted for the tweets in Armenian and/or Azerbaijani, the languages of the sides of the conflict, and/or in Turkish, Russian, Farsi, the languages of major neighboring countries, can provide more beneficial results, assuming that general population of those countries is not prevented from using Twitter freely. Especially a study in Armenian and/or Azerbaijani tweets may capture early indications of an upcoming rise in tension which we were able to observe in English tweets only before full scale war.

## Acknowledgments

We thank Emine Gülşen Torunoğlu Aslan for her assistance with the wording of this manuscript which greatly improved by her comments and reviews.